%% file: main_Arxiv.tex
\documentclass{article}

\usepackage{PRIMEarxiv}

\usepackage[utf8]{inputenc} 
\usepackage[T1]{fontenc}    
\usepackage{hyperref}       
\usepackage{url}            
\usepackage{booktabs}       
\usepackage{amsfonts}       
\usepackage{nicefrac}       
\usepackage{microtype}      
\usepackage{lipsum}
\usepackage{fancyhdr}       
\usepackage{graphicx}       
\graphicspath{{media/}}     

\usepackage{cite}
\usepackage{amsmath,amssymb,amsfonts}
\usepackage{graphicx}
\usepackage{textcomp}
\usepackage{xcolor}
\usepackage{tabularx}
\usepackage{caption}
\usepackage{algpseudocode}
\usepackage{algorithm}
\usepackage{float}

\usepackage{mdframed}

\input{pre}

\usepackage[affil-it]{authblk}  

\begin{document}

\pagestyle{fancy}
\thispagestyle{empty}
\rhead{ \textit{ }} 

\fancyhead[LO]{Running Title for Header}

\title{Geographical Information Alignment Boosts Traffic Analysis
 via Transpose Cross-attention
}

\author[1]{Xiangyu Jiang $^{1,}$\thanks{Corresponding author}    ,   Xiwen Chen} 
\author[1]{Hao Wang}
\author[1]{Abolfazl Razi}


\affil[1]{School of Computing, Clemson University, SC, USA}


\maketitle

\begin{abstract}
Traffic accident prediction is crucial for enhancing road safety and mitigating congestion, and recent Graph Neural Networks (GNNs) have shown promise in modeling the inherent graph-based traffic data. However, existing GNN-based approaches often overlook or do not explicitly exploit geographic position information, which often plays a critical role in understanding spatial dependencies. This is also aligned with our observation, where accident locations are often highly relevant. To address this issue, we propose a plug-in-and-play module for common GNN frameworks, termed Geographic Information Alignment (GIA). This module can efficiently fuse the node feature and geographic position information through a novel Transpose Cross-attention mechanism. Due to the large number of nodes for traffic data, the conventional cross-attention mechanism performing the node-wise alignment may be infeasible in computation-limited resources. Instead, we take the transpose operation for Query, Key, and Value in the Cross-attention mechanism, which substantially reduces the computation cost while maintaining sufficient information.
  Experimental results for both traffic occurrence prediction and severity prediction (severity levels based on the interval of recorded crash counts) on large-scale city-wise datasets confirm the effectiveness of our proposed method. For example, our method can obtain gains ranging from 1.3\% to 10.9\% in F1 score and 0.3\% to 4.8\% in AUC\footnote{This material is based upon the work supported by the National Science Foundation under Grant Number 2204721 and partially supported by our collaborative project with MIT Lincoln Lab under Grant Number 7000612889.}.
\end{abstract}


\section{Introduction}

\begin{figure}[!t]
		\centering
		\includegraphics[width=0.6\columnwidth]{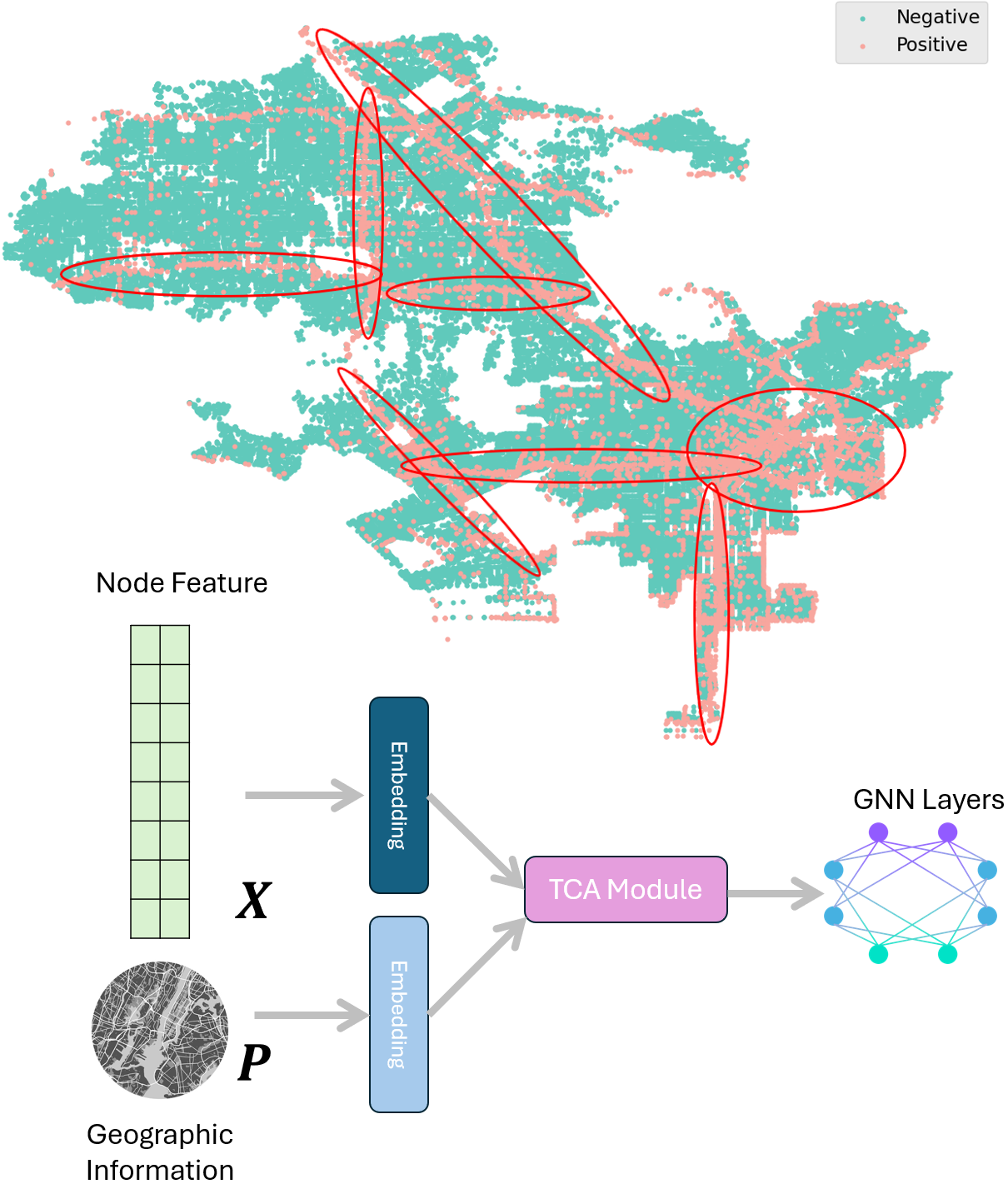}
		\caption{\textbf{Top:} Traffic Accident Map in Los Angeles shows a strong correlation between traffic accidents and location information. Positive denotes the crash/accident occurrence. We mark some exemplary accident clusters. \textbf{Bottom:} The proposed Geographic Information Alignment (GIA) through the Transpose Cross-attention (TCA) mechanism.}
        \captionsetup{justification=raggedright, singlelinecheck=false}
		\label{fig:Los_Angeles_Ground_truth}
	\end{figure}

	Road traffic accidents pose a significant global health and safety issue, as reported by the World Health Organization (WHO). The 2018 report estimated 1.35 million fatalities per year, and though this number slightly decreased to 1.19 million in the 2023 report, the issue remains critical \cite{who2018global,who2023global}. To mitigate the risks associated with road traffic incidents, predictive analytics have become a promising approach in modern traffic management systems. The importance of modeling traffic incident risks is well-recognized in the field of urban planning and public safety \cite{fadhel2024comprehensive,razi2023deep,bastola2024driving}.
 In the machine learning community, the use of traffic data is often large-scale, and it can be inherently possessed to a graph/network structure \cite{wang2022fast,rahmani2023graph,chen2022network}.
 For example, nodes represent physical locations, and edges represent the roads connecting them \cite{huang2023tap}.  
Therefore, learning graph-based representations of road networks enables understanding traffic flow and identifying high-risk areas for accidents. 
 It is noteworthy that graph-based data is often considered non-conventional due to its irregular structure with dynamics topology. To address this issue, several Graph Neural Networks (GNNs) are proposed \cite{kipf2016gcnconv,defferrard2016chebnet,hamilton2017inductive,shi2021graphtransformer}. Subsequently, many of the following works \cite{zhang2023fptn,zhao2023dynamic,lan2022dstagnn,chen2022bidirectional,huang2023tap} apply GNN to different traffic
tasks.

     \begin{figure*}[!t]
		\centering
		\includegraphics[width=0.95\textwidth]{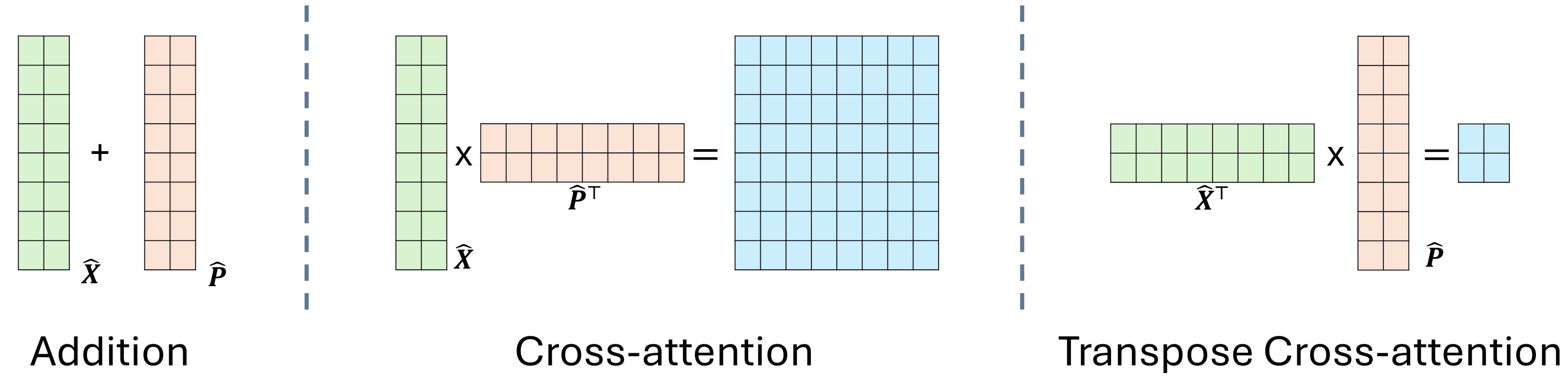}
		\caption{Comparison of different fusion strategies. \textbf{Left:} Simply take addition, \textbf{Middle:} The conventional Cross-attention, and \textbf{Right:} Our proposed Transpose Cross-attention (TCA) for feature-alignment. }
		\label{fig:fusion}
	\end{figure*}
 
	However, a general issue with GNNs is their permutation-invariant nature, which means they do not consider the order of nodes. In the context of traffic analysis, these models may neglect or do not explicitly exploit geographic position information.
 This can be problematic in traffic accident prediction because geographical context and the precise location of nodes are highly relevant. This is also aligned with our observation of real-world datasets \cite{huang2023tap}. In Fig. \ref{fig:Los_Angeles_Ground_truth}(\textbf{Left}), we can observe the accident distribution is often grouped according to some geographical patterns. Areas with dense traffic networks or critical intersections tend to experience higher accident frequencies, highlighting the importance of geographic context in traffic incident analysis.

    Therefore, we aim to explicitly harness this geographical information to GNN, and to this end, we propose a plug-in-and-play technique termed \textit{Geographical Information Alignment} (GIA) as shown in Fig. \ref{fig:Los_Angeles_Ground_truth}(\textbf{Right}), which can be integrated into general GNN frameworks. Intuitively, we can employ the Cross-attention mechanism to fuse the node feature and the geographical information. We note that the number of nodes $N$ often is substantial (e.g., over 10k), and directly applying Cross-attention requires $\mathcal{O}(N^2)$ complexity in memory and time, which can challenge the computation resources (In fact, we encountered out-of-memory issues when attempting this approach using one NVIDIA V100 GPU). Instead, we propose a \textbf{Transpose} Cross-attention mechanism, where we construct attention between each dimension and attempt to align features from edges and geographical information. This substantially reduces the complexity from $\mathcal{O}(N^2)$ to $\mathcal{O}(d^2)$, where $d$ denotes the number of dimensions and is much smaller than $N$ (e.g., $d=16$ in our experiments). Fig. \ref{fig:fusion} illustrates the different fusion strategies. The left one simply takes the addition of node features and position representation, which tends to limited learning performance. The middle one shows the conventional Cross-attention to fuse two features, which has a high computational cost, and the right one shows our proposed Transpose Cross-attention, which performs feature-wise alignment.
    
    Our analysis is based on a database of national traffic accident reports provided by \cite{huang2023tap}. We select some representative cities and states in our experiments, including Dallas, Houston, Miami, Orlando, Los Angeles, and New York. 
    We consider two important tasks involving accident occurrence prediction and severity prediction.
    The empirical results confirm the effectiveness of our enhanced GNN model, showing significant improvements in predictive performance across multiple state-of-the-art GNN models.

\section{RELATED WORK}
    \noindent\textbf{Graph Neural Networks in Traffic Analysis. }
    In current mainstream research, network, and graph data are widely used to deal with problems related to traffic flow \cite{10580972,10621553}. Therefore, many existing studies focus on network analysis~\cite{watts1998collective,boccaletti2014structure,barabasi1999emergence}.
    However, compared to common data types in the field of machine learning (such as images) \cite{10621553}, graph data are more difficult to process because their topological structure is often dynamic and large in scale~\cite{ding2019application}. To tackle this issue, the community developed and improved many methods, and the most popular paradigm is GNN. At a very early stage, spectral methods were introduced to perform convolution operations using the graph Laplacian operator~\cite{bruna2013spectral}. Afterward, Graph Convolutional Networks (GCN)~\cite{kipf2016gcn} significantly reduce the computational complexity to enhance its generalizability. Subsequently, many variants of GNN are developed with different modules \cite{defferrard2016chebnet,bianchi2021armanet,hamilton2017inductive,du2018tagcn}, such as attention mechanism \cite{velivckovic2017gat} or Transformer \cite{shi2021graphtransformer}.

	Graph Neural Networks (GNNs) have become increasingly important in traffic analysis due to their ability to model traffic networks as graphs, where nodes represent traffic entities like vehicles or road intersections, and edges represent interactions such as traffic flow or road connections. GNNs are particularly effective for tasks like traffic flow prediction, route optimization, and accident risk assessment. For example, recent models such as FPTN\cite{zhang2023fptn} enhance traffic forecasting by leveraging sensor-based data division and a Transformer encoder, which reduces computational demands while improving accuracy. Similarly, DyHSL\cite{zhao2023dynamic} utilizes a hypergraph neural network (HGNN) to capture dynamic and interactive spatio-temporal relationships, significantly improving forecasting performance across multiple datasets. DSTAGNN\cite{lan2022dstagnn} focuses on dynamically modeling spatial-temporal interactions in road networks, employing enhanced multi-head attention mechanisms and multi-scale gated convolutions for more precise traffic predictions. Moreover, Bi-STAT\cite{chen2022bidirectional} advances traffic forecasting by using adaptive spatial-temporal transformers, effectively handling diverse tasks and leveraging historical data for more accurate predictions. Finally, STFGNN~\cite{li2021spatial} integrates data-driven temporal and spatial graphs and gated convolutions, successfully managing long sequences of traffic data, making it a robust model for traffic forecasting.

        Our work is based on \cite{huang2023tap} for the large-scale traffic accident analysis with GNNs. In this work, the authors investigate a series of recent state-of-the-art GNN models (such as \cite{kipf2016gcnconv,defferrard2016chebnet,hamilton2017inductive,shi2021graphtransformer}) and further propose a new GNN-based method to analyze the traffic accident occurrence and severity. 
	We reiterate that a significant limitation of these approaches is their reliance on topological features of the graph while often ignoring or not explicitly employing the absolute spatial coordinates of the nodes, making it difficult to fully exploit prior to the geographic information. Accordingly, this may limit the models' predictive power for tasks such as traffic accident prediction. Hence, in this work, we propose a method to align position information with node features to mitigate this concern.

    \noindent\textbf{Positional Encoding.}
The proposed method is relevant to positional encoding techniques; however, we realize that there exists an essential difference between them.
	Position encoding has been extensively studied in Natural Language Processing (NLP), particularly in the context of the Transformer architecture~\cite{vaswani2017attention}, where sinusoidal position encoding is used to help models capture the sequential order of words. This technique is unlearnable, which often leads to limited performance. In the domain of GNNs, researchers have started to investigate methods to encode positional information within graphs. A known work presented in ~\cite{dwivedi2023benchmarking} introduced positional encoding for GNN. However, similar to sinusoidal position encoding, they directly add position information and latent features, which apparently degenerate learning ability. Instead, we propose the Transpose Cross-attention to better and efficiently integrate position and node features.
 In traffic analysis, although some recent studies have experimented with incorporating node coordinates into GNNs by injecting the coordinate information as node features~\cite{zhou2023Spatiotemporal}, these efforts often lack a systematic approach to position encoding. In contrast, our method offers a universal strategy to elegantly impose the position information to GNNs.

	\section{Method }
	In this section, we present the problem formulation for traffic accident prediction, followed by our proposed method, the \textit{Geographical Information Alignment} (GIA) through a novel Transpose Cross-attention mechanism.
	\subsection{Problem Formulation}
	Following \cite{huang2023tap}, the task of traffic accident prediction is formulated as a node-wise classification problem in a monolithic graph under the popular transductive setup\cite{prince2023understanding}. 
In this context, each dataset (i.e., a city) is presented as a large graph with $N$ nodes and $E$ edges. Subsequently, this graph can be presented by three matrices $\boldsymbol{A}\in\mathbb{R}^{N\times N}$, $\boldsymbol{X}\in\mathbb{R}^{ N\times D_1}$, and $\boldsymbol{E}\in\mathbb{R}^{ E\times D_2}$, denoting the adjacency matrix, node embedding, and edge embedding, respectively. Here, $D_1$ and $D_2$ denote the number of dimensions of node and edge features, respectively. We also have partial node labels, with the index set of these labeled nodes denoted as $\mathcal{V}_{\text{train}}$. The set of nodes whose labels are unknown and need to be predicted is denoted as $\mathcal{V}_{\text{test}}$. 
	The goal is to classify nodes based on accident occurrence or severity. Accident occurrence prediction is a binary classification, indicating whether an accident has occurred at a node or not. In contrast, the severity prediction involves eight categories, where each category is defined based on an interval of recorded crash counts. Please refer to  \cite{huang2023tap} for more detail.

	\subsection{Proposed Method}
    Here, we delve into the proposed \textit{Geographical Information Alignment} and demonstrate how to incorporate it with general GNN models.

	\subsubsection{Position Encoding}
 The goal is to encode the semantic information of a node's absolute positions into its feature representation. First, we can obtain the embedding of the node features as  $\hat{\boldsymbol{X}}  \leftarrow f_{\texttt{embed}}(\boldsymbol{X} )\in\mathbb{R}^{ N\times D_n}$. Here $N$ and $D_n$ denote the number of nodes and the dimensions of latent space, respectively. This is because, in the raw data space, different features represent different physical meanings (e.g., junction types, speed limit, etc,. ). Directly imposing position information in that space may constrain the learning ability. We also want to mention that the embedding layer $f_{\texttt{embed}}(\cdot)$ comes from the original architecture of a GNN, and we do not intend to modify it.  Given a geographical position $\boldsymbol{P}\in\mathbb{R}^{N\times 2}$, we then map it to a latent space via a linear layer to match the dimensions of the input edge feature, which is denoted as $\hat{\boldsymbol{P}}\leftarrow f_{\texttt{linear}}(\boldsymbol{P})\in\mathbb{R}^{ N\times D_n}$. Subsequently, we consider to fuse it with $\hat{\boldsymbol{P}}\in\mathbb{R}^{ N\times D_n}$ and $\hat{\boldsymbol{X}}\in\mathbb{R}^{ N\times D_n}$. An intuitive solution is employing the cross-attention mechanism to perform node-wise alignment:

 \begin{align}
     \boldsymbol{H} = \operatorname{softmax}\underbrace{\left(\frac{\hat{\boldsymbol{X}} \hat{\boldsymbol{P}}^\top}{\sqrt{D_n}}\right)}_{N\times N} ~\hat{\boldsymbol{P}}.
 \end{align}

Following the common operation, we can apply the respective linear layers for the Query, Key, and Value. However, the main concern here is that in our graph-based problem, the number of nodes is often very large, and we have $N>>D_n$, which leads to the computation and memory cost for the attention matrix requiring an $\mathcal{O}(n^2)$. This may be unacceptable for a resource-constrained platform. 

To address the concern, we propose the \textit{\textbf{Transpose} Cross-attention} mechanism. Instead of performing the node-wise alignment, we prefer feature-wise alignment. Mathematically,
\begin{mdframed}
 \begin{align}
     \boldsymbol{H} = \operatorname{softmax}\underbrace{\left(\frac{\hat{\boldsymbol{X}^{\textcolor[rgb]{0.0, 0.0, 0.9}{\top}}} \hat{\boldsymbol{P}}}{\sqrt{N}}\right)}_{D_n \times D_n} ~\hat{\boldsymbol{P}}^{\textcolor[rgb]{0.0, 0.0, 0.9}{\top}}.
 \end{align}
  \end{mdframed}
This substantially reduces the complexity from $\mathcal{O}(n^2)$ to $\mathcal{O}(D_n^2)$. Another justification is this operation can also contain sufficient information from the perspective of linear algebra. This is because $rank(\hat{\boldsymbol{X}} \hat{\boldsymbol{P}}^\top) = rank(\hat{\boldsymbol{X}}^\top \hat{\boldsymbol{P}}) = \min(N,D_n)=D_n$. We also added a residual term to stabilize the training as,
\begin{align}\label{eq:gia-full}
    \hat{\boldsymbol{X}}\leftarrow f_{\texttt{linear}}({\boldsymbol{X}}) +  \operatorname{softmax}\left(\frac{\hat{\boldsymbol{X}^{\textcolor[rgb]{0.0, 0.0, 0.9}{\top}}} \hat{\boldsymbol{P}}}{\sqrt{N}}\right) \hat{\boldsymbol{P}}^{\textcolor[rgb]{0.0, 0.0, 0.9}{\top}}.
\end{align}
The overall architecture is presented in Fig. \ref{fig:Los_Angeles_Ground_truth}.

   \begin{figure*}[hbtp]
		\centering
		\includegraphics[width=0.99\textwidth]{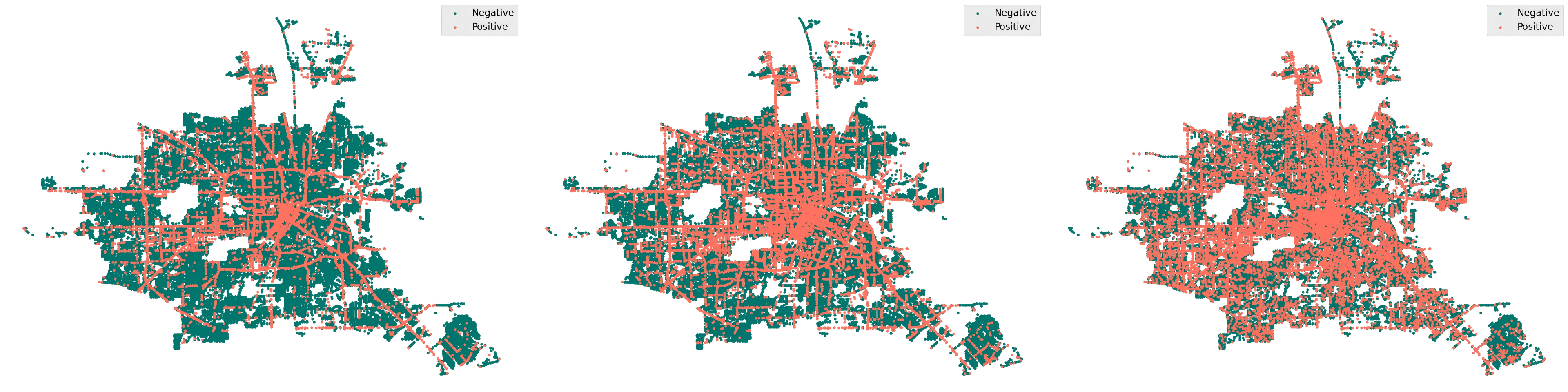}
		\caption{Comparison of traffic accident predictions in the Houston area using the GEN model(left), GEN-GIA model(middle), and the ground truth(right). The GEN-GIA model shows improved accuracy in predicting high-accident areas by integrating positional encoding.}
		
		\label{fig:houston_comp}
	\end{figure*}

	\section{Experiments}
	
\noindent\textbf{Setup.} In this section, we evaluate the performance of the proposed GNN-based model variants for traffic incident prediction. We employ multiple model baselines and analyze their performance on large-scale traffic datasets provided by \cite{huang2023tap}. 
 We select representative cities and states, such as Miami, Los Angeles and New York to ensure geographical diversity and capture a wide range of traffic patterns. These datasets vary in size and structure, providing a comprehensive testbed for traffic analysis. Detailed information on the number of nodes and edges for each dataset can be found in Table \ref{tab:datasets}.  
 It is noteworthy to mention that the datasets are very unbalanced, which challenges most deep learning or machine learning models. 

\begin{table}[h]
    \centering

    
    \resizebox{0.8\columnwidth}{!}{%
        \begin{tabular}{|c|c|c|c|c|c|c|}
            \hline
            \textbf{Dataset} & \textbf{\begin{tabular}[c]{@{}c@{}}Miami \\ (FL)\end{tabular}} & \textbf{\begin{tabular}[c]{@{}c@{}}Los Angeles\\ (CA)\end{tabular}} & \textbf{\begin{tabular}[c]{@{}c@{}}Orlando \\ (FL)\end{tabular}} & \textbf{\begin{tabular}[c]{@{}c@{}}Dallas\\ (TX)\end{tabular}} & \textbf{\begin{tabular}[c]{@{}c@{}}Houston\\ (TX)\end{tabular}} & \textbf{\begin{tabular}[c]{@{}c@{}}New York\\ (NY)\end{tabular}} \\ \hline
            \textbf{\# of Nodes} & 8461 & 49251 & 7513 & 36150 & 59711 & 55404 \\ \hline
            \textbf{\# of Edges} & 22648 & 135547 & 18216 & 92348 & 148937 & 140005 \\ \hline
        \end{tabular}%
    }
        \caption{Summary of the datasets used in the experiments.}
        \label{tab:datasets}
\end{table}

	The dataset is stratified into training (60\%), validation (20\%), and test (20\%) sets, with stratification based on the incident labels. All features are normalized using the MinMaxScaler for each split.

	\noindent\textbf{Baselines.} To validate that our proposed method is model-agnostic, we choose 12 different state-of-the-art GNN models, including (1) \textbf{GCN}: Graph Convolutional Networks \cite{kipf2016gcnconv}, (2) \textbf{ChebNet}: graph convolutional networks with spetral filtering \cite{defferrard2016chebnet}, (3) \textbf{ARMANet}: Graph neural networks with convolutional auto-regressive moving average (ARMA) filters \cite{bianchi2021armanet}. (4) \textbf{GraphSAGE}: inductive representation learning \cite{hamilton2017inductive}, (5) \textbf{TAGCN}: Topology adaptive graph convolutional networks \cite{du2018tagcn}, (6) 
	\textbf{GIN}: Graph Isomorphism Networks \cite{xu2018gnn}, (7) \textbf{GAT}: Graph attention networks \cite{velivckovic2017gat}, (8) \textbf{MPNN}: Message Passing Neural Network \cite{gilmer2017mpnn}, (9) \textbf{CGC}: Crystal graph convolutional neural network \cite{xie2018cgc}, (10) \textbf{GEN}: GENeralized graph convolutional neural networks \cite{li2020gen}, (11) \textbf{Graphformer}: Graph transformers \cite{shi2021graphtransformer}, and (12) \textbf{TRAVEL}: the recent GNN model proposed for our problem \cite{huang2023tap}. For the sake of fair comparison, Each model, including the baselines and our proposed variants, applies a unified training regimen, where we adjust the training epochs of 500 to make the model fully converge. The other hyperparameters follow the default setting in \cite{huang2023tap}.

	\noindent\textbf{Main Results.} 
 The main results are shown in Table \ref{tab:city-performance}. The primary observation is that by the proposed GIA, all GNN models demonstrate significant improvements across different cities. Specifically, in the occurrence prediction task (Table \ref{tab:city-performance}), GNN models achieve a  {1.53\% to 10.9\%} gain in  {F1 score} and a  {1.26\% to 4.8\%} gain in  {AUC} when incorporating our proposed GIA. For instance, models like  {GCN-GIA},  {ChebNet-GIA}, and  {GIN-GIA} demonstrate F1 score improvements of 1.53\%, 2.6\%, and 4.51\%, respectively. The model  {GIN-GIA} shows an exceptional improvement in the city of Miami, achieving a remarkable  {10.9\% increase in F1 score}(see Fig. \ref{fig:houston_comp}) and a  {4.0\% increase in AUC}.

	More importantly, the  {TRAVEL-GIA} model, which previously had the best performance, achieved an additional  {1.74\% gain in F1 score} and a  {1.26\% gain in AUC} compared to its non-PE variant. Likewise, in  {GEN-GIA}, the F1 score shows a  {4.69\% improvement}, with notable enhancements in cities like Los Angeles and Dallas. These results underscore the critical role of incorporating geographic positional information to boost GNN performance on traffic prediction tasks.
	
	Similarly, in severity prediction tasks (Table \ref{tab:severity-performance}), all GNN models experience consistent improvements. In this task, due to the already high performance of the original models (around 85\%), the improvements are relatively modest but still significant. Notably,  {Transformer-GIA} achieves the highest F1 score of  {86.59\%}, with an improvement of  {0.79\%}. The  {CGC-GIA} model also exhibits strong performance, reaching an F1 score of  {86.48\%} with a  {0.68\%} improvement.  {TRAVEL-GIA} shows the largest relative improvement of  {1.12\%}, reaching an F1 score of  {85.82\%}. These results confirm that integrating GIA leads to measurable performance gains, even in tasks where baseline models already perform well.

\begin{table*}[t]
		\resizebox{\textwidth}{!}{%
			\begin{tabular}{lcccccccccccccc}
				\hline
				Dataset & \multicolumn{2}{c}{Mean} & \multicolumn{2}{c}{Miami (FL)} & \multicolumn{2}{c}{Los Angeles(CA)} & \multicolumn{2}{c}{Orlando (FL)} & \multicolumn{2}{c}{Dallas (TX)} & \multicolumn{2}{c}{Houston (TX)} & \multicolumn{2}{c}{New York (NY)} \\ \hline
				Method & \multicolumn{1}{l}{F1} & \multicolumn{1}{l}{AUC} & \multicolumn{1}{l}{F1} & \multicolumn{1}{l}{AUC} & \multicolumn{1}{l}{F1} & \multicolumn{1}{l}{AUC} & \multicolumn{1}{l}{F1} & \multicolumn{1}{l}{AUC} & \multicolumn{1}{l}{F1} & \multicolumn{1}{l}{AUC} & \multicolumn{1}{l}{F1} & \multicolumn{1}{l}{AUC} & \multicolumn{1}{l}{F1} & \multicolumn{1}{l}{AUC} \\
				\hline
				GCN & 34.53 & 72.88 & 20.0±3.3 & 68.5±3.3 & 40.2±1.1 & 80.4±0.3 & 51.6±0.8 & 73.1±1.2 & 39.8±1.9 & 73.1±0.4 & 16.4±1.3 & 66.7±0.2 & 39.2±3.7 & 75.5±0.4 \\
				GCN-GIA & 36.06 & 75.54 & 21.73±2.77 & 72.69±0.18 & 40.74±0.47 & 82.79±0.15 & 52.32±0.18 & 76.65±0.16 & 41.16±0.80 & 74.00±0.11 & 18.83±1.47 & 67.57±0.20 & 41.57±0.91 & 79.52±0.13 \\
				\rowcolor[HTML]{72aec4} 
				$\Delta$ & 1.53 & 2.66 & 1.7 & 4.1 & 0.5 & 2.3 & 0.7 & 3.5 & 1.3 & 0.9 & 2.4 & 0.8 & 2.3 & 4 \\ \hline
				ChebNet & 36.72 & 75.45 & 20.7±2.9 & 71.3±3.6 & 39.8±1.8 & 81.0±0.3 & 53.1±0.6 & 76.7±1.6 & 42.0±0.5 & 75.8±0.4 & 23.8±0.5 & 69.6±0.5 & 40.9±4.3 & 78.3±1.1 \\
				ChebNet-GIA & 39.32 & 77.29 & 25.80±0.83 & 73.16±0.87 & 42.43±1.35 & 83.86±0.33 & 55.52±1.50 & 78.16±0.63 & 43.46±0.07 & 76.64±0.08 & 25.30±0.96 & 70.26±0.27 & 43.41±0.74 & 81.68±0.16 \\
				\rowcolor[HTML]{72aec4} 
				$\Delta$ & 2.6 & 1.84 & 5.1 & 1.8 & 2.6 & 2.8 & 2.4 & 1.4 & 1.4 & 0.8 & 1.5 & 0.6 & 2.5 & 3.3 \\ \hline
				ARMANet & 36.37 & 74.77 & 19.2±3.3 & 69.5±3.5 & 40.8±1.0 & 80.9±0.4 & 51.5±1.3 & 75.7±1.4 & 41.2±0.5 & 75.6±0.2 & 23.1±0.4 & 69.2±0.7 & 42.4±1.1 & 77.7±0.6 \\
				ARMANet-GIA & 39.51 & 77.25 & 25.69±0.29 & 74.32±0.06 & 43.70±1.30 & 83.55±0.35 & 57.43±0.20 & 78.44±0.45 & 43.66±1.13 & 76.23±0.14 & 24.53±0.69 & 69.90±0.34 & 42.03±0.78 & 81.04±0.42 \\
				\rowcolor[HTML]{72aec4} 
				$\Delta$ & 3.14 & 2.48 & 6.4 & 4.8 & 2.9 & 2.6 & 5.9 & 2.7 & 2.4 & 0.6 & 1.4 & 0.7 & -0.4 & 3.3 \\ \hline
				GraphSAGE & 37.55 & 73.57 & 20.7±2.4 & 67.6±2.8 & 41.6±0.5 & 80.5±0.3 & 52.6±1.3 & 74.1±1.2 & 44.2±0.5 & 74.4±0.3 & 23.7±0.4 & 68.5±0.4 & 42.5±1.1 & 76.3±0.1 \\
				GraphSAGE-GIA & 38.94 & 75.98 & 25.44±3.05 & 71.41±1.65 & 44.05±1.31 & 82.86±0.68 & 54.51±1.00 & 77.00±0.64 & 42.49±0.66 & 75.16±0.12 & 24.50±0.41 & 69.74±0.53 & 42.68±0.94 & 79.68±0.01 \\
				\rowcolor[HTML]{72aec4} 
				$\Delta$ & 1.39 & 2.41 & 4.7 & 3.8 & 2.4 & 2.3 & 1.9 & 2.9 & -1.8 & 0.7 & 0.8 & 1.2 & 0.1 & 3.3 \\ \hline
				TAGCN & 39.85 & 77.40 & 25.2±1.1 & 73.5±2.4 & 49.5±0.7 & 84.7±0.2 & 53.3±2.5 & 77.2±1.2 & 45.4±0.4 & 77.0±0.5 & 23.7±0.6 & 70.5±0.3 & 42.0±1.1 & 81.5±0.2 \\
				TAGCN-GIA & 42.77 & 79.11 & 30.43±1.68 & 75.64±0.75 & 53.35±1.78 & 86.24±0.21 & 57.87±0.62 & 80.13±0.46 & 47.38±0.65 & 77.66±0.01 & 25.20±0.92 & 70.95±0.26 & 42.40±0.60 & 84.06±0.30 \\
				\rowcolor[HTML]{72aec4} 
				$\Delta$ & 2.92 & 1.71 & 5.2 & 2.1 & 3.8 & 1.5 & 4.5 & 2.9 & 1.9 & 0.6 & 1.5 & 0.4 & 0.4 & 2.5 \\ \hline
				GIN & 37.17 & 75.57 & 22.8±1.2 & 72.7±2.6 & 41.6±0.7 & 81.8±0.2 & 54.7±1.4 & 76.6±1.1 & 41.3±2.0 & 75.2±0.3 & 20.9±1.0 & 68.0±0.3 & 41.7±2.1 & 79.1±0.5 \\
				GIN-GIA & 41.68 & 77.94 & 33.77±0.15 & 76.71±3.35 & 46.51±0.54 & 84.37±0.49 & 57.81±1.27 & 78.94±0.08 & 43.81±0.61 & 76.93±0.35 & 24.50±1.49 & 68.78±0.07 & 43.66±0.47 & 81.92±0.28 \\
				\rowcolor[HTML]{72aec4} 
				$\Delta$ & 4.51 & 2.37 & 10.9 & 4 & 4.9 & 2.5 & 3.1 & 2.3 & 2.5 & 1.7 & 3.6 & 0.7 & 1.9 & 2.8 \\ \hline
				GAT & 36.93 & 73.47 & 22.6±1.5 & 68.3±3.0 & 41.6±0.4 & 80.9±0.2 & 55.3±1.3 & 74.1±1.0 & 42.1±1.5 & 73.6±0.3 & 17.8±0.8 & 67.3±0.3 & 42.2±0.5 & 76.6±0.4 \\
				GAT-GIA & 39.13 & 76.11 & 27.69±2.23 & 72.03±2.31 & 45.28±1.24 & 83.92±0.49 & 56.55±1.11 & 77.59±0.08 & 42.01±0.05 & 74.59±0.10 & 20.91±1.15 & 68.28±0.33 & 42.37±0.03 & 80.24±0.03 \\
				\rowcolor[HTML]{72aec4} 
				$\Delta$ & 2.2 & 2.64 & 5 & 3.7 & 3.6 & 3 & 1.2 & 3.4 & -0.1 & 0.9 & 3.1 & 0.9 & 0.1 & 3.6 \\ \hline
				MPNN & 44.63 & 81.32 & 38.8±2.1 & 82.4±1.0 & 46.0±1.6 & 83.9±0.2 & 61.4±2.5 & 81.8±0.7 & 48.5±1.9 & 79.4±0.4 & 28.2±1.7 & 73.5±0.5 & 44.9±0.8 & 86.9±0.4 \\
				MPNN-GIA & 47.01 & 82.96 & 43.03±1.44 & 84.72±0.38 & 48.88±0.98 & 86.91±0.51 & 63.72±1.05 & 82.45±0.95 & 50.23±0.73 & 81.26±0.60 & 30.72±0.43 & 74.61±0.26 & 45.49±0.08 & 87.84±0.19 \\
				\rowcolor[HTML]{72aec4} 
				$\Delta$ & 2.38 & 1.64 & 4.2 & 2.3 & 2.8 & 3 & 2.3 & 0.6 & 1.7 & 1.8 & 2.5 & 1.1 & 0.5 & 0.9 \\ \hline
				CGC & 42.47 & 79.83 & 34.4±2.7 & 79.5±1.5 & 45.0±1.2 & 81.5±0.2 & 59.0±2.1 & 81.1±0.8 & 48.5±0.5 & 79.2±0.7 & 27.3±1.9 & 72.3±0.1 & 40.6±1.2 & 85.4±0.8 \\
				CGC-GIA & 46.08 & 82.29 & 38.33±3.06 & 84.16±0.26 & 47.98±1.94 & 85.88±0.57 & 63.12±0.60 & 82.74±0.47 & 51.20±0.50 & 80.37±0.33 & 33.02±2.86 & 73.13±0.13 & 42.83±0.15 & 87.47±0.25 \\
				\rowcolor[HTML]{72aec4} 
				$\Delta$ & 3.61 & 2.46 & 3.9 & 4.6 & 2.9 & 4.3 & 4.1 & 1.6 & 2.7 & 1.1 & 5.7 & 0.8 & 2.2 & 2 \\ \hline
				Graphformer & 45.13 & 81.32 & 37.7±3.3 & 81.0±1.9 & 48.9±0.3 & 83.8±0.3 & 62.9±1.6 & 82.0±0.7 & 49.8±0.7 & 80.0±0.7 & 28.4±0.7 & 73.9±0.4 & 43.1±0.7 & 87.2±0.4 \\
				Transformer-GIA & 48.02 & 83.17 & 42.04±1.35 & 85.04±0.47 & 51.74±1.45 & 86.59±0.51 & 64.52±0.83 & 83.02±0.63 & 51.33±0.80 & 80.96±0.21 & 33.22±0.58 & 74.83±0.24 & 45.26±0.74 & 88.56±0.16 \\
				\rowcolor[HTML]{72aec4} 
				$\Delta$ & 2.89 & 1.85 & 4.3 & 4 & 2.8 & 2.7 & 1.6 & 1 & 1.5 & 0.9 & 4.8 & 0.9 & 2.1 & 1.3 \\ \hline
				GEN & 49.07 & 80.97 & 44.9±3.1 & 81.0±2.4 & 48.6±6.2 & 82.7±0.9 & 63.0±1.1 & 81.2±0.9 & 56.5±1.7 & 79.5±0.1 & 34.1±6.0 & 73.7±0.4 & 47.3±1.4 & 87.7±0.9 \\
				GEN-GIA & 53.76 & 83.62 & 51.96±0.66 & 86.14±0.55 & 55.44±0.17 & 87.67±0.22 & 65.42±1.04 & 82.79±1.08 & 56.14±0.99 & 81.06±0.49 & 42.95±0.94 & 75.35±0.89 & 50.65±2.37 & 88.70±0.07 \\
				\rowcolor[HTML]{72aec4} 
				$\Delta$ & 4.69 & 2.65 & 7 & 5.1 & 6.8 & 4.9 & 2.4 & 1.5 & -0.4 & 1.5 & 8.8 & 1.6 & 3.3 & 1 \\ \hline
				TRAVEL & 54.62 & 82.77 & 51.9±1.0 & 84.9±0.9 & 55.3±0.9 & 85.9±0.5 & 65.0±0.4 & 82.3±0.4 & 58.0±0.9 & 80.8±0.7 & 46.4±0.7 & 74.5±0.3 & 51.1±0.9 & 88.2±0.2 \\
				TRAVEL-GIA & 56.36 & 84.03 & 54.91±1.04 & 86.44±0.41 & 58.07±0.47 & 87.99±0.29 & 65.71±0.21 & 83.58±0.13 & 58.29±0.12 & 81.59±0.22 & 47.34±0.66 & 75.79±0.58 & 53.83±0.95 & 88.76±0.05 \\
				\rowcolor[HTML]{72aec4} 
				$\Delta$ & 1.74 & 1.26 & 3 & 1.5 & 2.7 & 2 & 0.7 & 1.2 & 0.2 & 0.7 & 0.9 & 1.2 & 2.7 & 0.5 \\ \hline
			\end{tabular}%
		}
		\caption{ City-wise accident occurrence prediction results in terms of F1 score and AUC. $\Delta$ denotes the gain obtained by imposing our proposed Geographical Information Alignment(with the suffix "-GIA") in the neural networks.}
        \captionsetup{justification=raggedright, singlelinecheck=false}
		\label{tab:city-performance}
	\end{table*}

\begin{table}[!h]
	\centering
    \vspace{1em}
        \label{table:aba}
        \resizebox{0.4\linewidth}{!}{
	\begin{tabular}{|l|c|c|}
		\hline
		{Model}     & {F1  (\%)} & \(\Delta {F1} \% \) \\ \hline
		{Transformer-GIA}  & 86.59                            & +0.79                    \\
		{CGC-GIA}   & 86.48                            & +0.68                    \\
		{TRAVEL-GIA} & 85.82            & +1.12          \\ \hline
	\end{tabular}
 }
 	\caption{Comparison of different models on Severity Prediction tasks.}
	\label{tab:severity-performance}
\end{table}
 
In conclusion, the addition of GIA provides a consistent performance boost across multiple models in the severity prediction task despite the already high baseline performance. This demonstrates that GIA is an effective enhancement for traffic incident severity prediction, capable of improving model performance even when the original models are already achieving strong results. The observed improvements, while modest in absolute terms, are significant given the high initial performance levels and highlight the potential of GIA in refining state-of-the-art models for this critical task.

    \begin{table}[!h]
	\centering
\resizebox{0.6\linewidth}{!}{
\begin{tabular}{lcccccc}
 \hline
\multicolumn{1}{|c|}{\multirow{2}{*}{Model}} & \multicolumn{2}{c|}{ARMANet} & \multicolumn{2}{c|}{GIN} & \multicolumn{2}{c|}{GEN} \\
\multicolumn{1}{|c|}{} & F1 & \multicolumn{1}{c|}{$\Delta$ F1} & F1 & \multicolumn{1}{c|}{$\Delta$ F1} & F1 & \multicolumn{1}{c|}{$\Delta$ F1} \\ \hline
\multicolumn{1}{|l|}{Baseline} & 36.37 & \multicolumn{1}{c|}{-} & 49.07 & \multicolumn{1}{c|}{-} & 49.07 & \multicolumn{1}{c|}{-} \\
\multicolumn{1}{|l|}{+GIA w/o TCA } & 38.81 & \multicolumn{1}{c|}{+2.44} & 38.49 & \multicolumn{1}{c|}{+1.32} & 51.59 & \multicolumn{1}{c|}{+2.52} \\

\multicolumn{1}{|l|}{+GIA} & \textbf{39.63} & \multicolumn{1}{c|}{\textbf{+3.26}} & \textbf{40.94} & \multicolumn{1}{c|}{\textbf{+3.77}} & \textbf{53.97} & \multicolumn{1}{c|}{\textbf{+4.9}} \\ 
\hline
\multicolumn{1}{|l|}{Sinusoidal} & 38.28 & \multicolumn{1}{c|}{+1.91} & 38.33 & \multicolumn{1}{c|}{+1.16} & 51.05 & \multicolumn{1}{c|}{+1.98} \\

\hline
\end{tabular}%
    
}
	 \caption{Ablation study comparing the performance of different GNN variants on accident occurrence prediction tasks with and without positional encoding techniques. The methods include the original GNN models (baseline), models without TCA (only linear residual term), models with Sinusoidal Encoding, and the full model.
 }
 \label{table:aba}
\end{table}

\noindent\textbf{Ablation Study.} Table \ref{table:aba} presents the results of our ablation study, where we compare the performance of different variations of GNN models across three methods: ARMANet, GIN, and GEN.
The first row corresponds to the original model without any positional encoding. The second
row shows we only keep the linear residual term (i.e., the first term presented in Eq.\ref{eq:gia-full}). This can also be viewed as a simple way for linear positional encoding. Finally, we will add the most popular Sinusoidal PE for comparison.
As shown in Table \ref{table:aba}, both linear and Sinusoidal positional encodings provide improvements over the original models, with Linear Encoding leading to a slightly higher improvement than Sinusoidal Encoding across all three models. This may be because it is learnable. Finally, we show the full model, demonstrating the performance achieved when Geographic Interaction Attention (GIA) is integrated into the system. This yields the highest improvements, with F1 scores increasing by 3.26\%, 3.77\%, and 4.9\% for ARMANet, GIN, and GEN, respectively, demonstrating the effectiveness of each component in our design for these traffic analysis tasks.

For the severity prediction tasks (Table \ref{table:aba_sev}), we observe a similar trend, with GIA once again providing the largest performance boost. Although the baseline models already perform well, GIA enhances F1 scores by 0.95\% for ARMANet, 0.64\% for GIN, and 0.75\% for GEN, confirming its positive impact even in high-performing models.

    \begin{table}[!t]
	\centering

\resizebox{0.6\linewidth}{!}{
\begin{tabular}{lcccccc}
 \hline
\multicolumn{1}{|c|}{\multirow{2}{*}{Model}} & \multicolumn{2}{c|}{ARMANet} & \multicolumn{2}{c|}{GIN} & \multicolumn{2}{c|}{GEN} \\
\multicolumn{1}{|c|}{} & F1 & \multicolumn{1}{c|}{$\Delta$ F1} & F1 & \multicolumn{1}{c|}{$\Delta$ F1} & F1 & \multicolumn{1}{c|}{$\Delta$ F1} \\ \hline
\multicolumn{1}{|l|}{Baseline} & 81.4 & \multicolumn{1}{c|}{-} & 81.3 & \multicolumn{1}{c|}{-} & 83.6 & \multicolumn{1}{c|}{-} \\

\multicolumn{1}{|l|}{+GIA w/o TCA } & 81.42 & \multicolumn{1}{c|}{+0.02} & 81.58 & \multicolumn{1}{c|}{+0.28} & 83.67 & \multicolumn{1}{c|}{+0.07} \\

\multicolumn{1}{|l|}{+GIA} & \textbf{82.35} & \multicolumn{1}{c|}{\textbf{+0.95}} & \textbf{81.94} & \multicolumn{1}{c|}{\textbf{+0.64}} & \textbf{84.35} & 
\multicolumn{1}{c|}{\textbf{+0.75}} \\ 
\hline

\multicolumn{1}{|l|}{Sinusoidal} & 81.5 & \multicolumn{1}{c|}{+0.1} & 81.41 & \multicolumn{1}{c|}{+0.11} & 83.89 & \multicolumn{1}{c|}{+0.29} \\

\hline
\end{tabular}%
    
}
	 \caption{Ablation study comparing the performance of different GNN variants on severity prediction tasks with and without positional encoding techniques. The methods include the original GNN models (baseline), models without TCA (only linear residual term), models with Sinusoidal Encoding, and the full GIA model.}
  \label{table:aba_sev}
\end{table}
	\section{Conclusion}
	In this paper, we aim to address the common limitation of GNNs, which often fail or explicitly incorporate spatial information in traffic prediction tasks. To this end, we propose a Geographical Information Alignment (GIA) module to enhance GNNs for addressing this constraint. Specifically, the GIA effectively integrates geographic positional information with node features using the proposed Transpose Cross-attention mechanism with a substantially low computation overhead. Our extensive experiments, conducted on large-scale traffic datasets from various cities, demonstrate significant performance improvements across multiple state-of-the-art GNN models, both in accident occurrence and severity prediction tasks. For example, our method can obtain up to 10.9\% and 4.8\% gain in F1 score and AUC, respectively.

\section*{Acknowledgments}
This material is based upon the work supported by the National Science Foundation under Grant Number 2204721 and partially supported by our collaborative project with MIT Lincoln Lab under Grant Number 7000612889.

\bibliographystyle{unsrt}  
\bibliography{reference}

\end{document}

%% file: pre.tex
\usepackage[utf8]{inputenc} 
\usepackage[T1]{fontenc}    
\usepackage{hyperref}       
\usepackage{url}            
\usepackage{booktabs}       
\usepackage{amsfonts}       
\usepackage{nicefrac}       
\usepackage{microtype}      

\usepackage[utf8]{inputenc} 
\usepackage[T1]{fontenc}    
\usepackage{hyperref}       

\usepackage{url}            
\usepackage{booktabs}       
\usepackage{amsfonts}       
\usepackage{nicefrac}       
\usepackage{microtype}      
\usepackage{lipsum}
\usepackage{graphicx}       
\graphicspath{{media/}}     
\usepackage{mathtools}
\usepackage{amsmath,amssymb,amsfonts}

\usepackage{graphicx}
\usepackage{textcomp}
\usepackage{xcolor}
\usepackage{comment}
\usepackage[T1]{fontenc}
\usepackage[utf8]{inputenc}
\usepackage{verbatim}
\usepackage{soul} 
\usepackage[hyphenbreaks]{breakurl}
 \usepackage{float}
\usepackage{hyperref}
\usepackage[hyphenbreaks]{breakurl}
\usepackage{multirow}
\usepackage{adjustbox}

\usepackage{enumerate}
\usepackage{colortbl}
\usepackage{booktabs}
\usepackage{array}
\usepackage[english]{babel}
\usepackage{amsthm}
\usepackage{setspace}
\theoremstyle{plain}

\theoremstyle{definition}

\theoremstyle{remark}

\usepackage{tikz}
\usetikzlibrary{shadows}

%% file: main_Arxiv.bbl
\begin{thebibliography}{10}

\bibitem{who2018global}
WorldHealthOrganization.
\newblock Global status report on road safety 2018, 2018.

\bibitem{who2023global}
WorldHealthOrganization.
\newblock Global status report on road safety 2023, 2023.

\bibitem{fadhel2024comprehensive}
Mohammed~A Fadhel, Ali~M Duhaim, Ahmed Saihood, Ahmed Sewify, Mokhaled~NA Al-Hamadani, AS~Albahri, Laith Alzubaidi, Ashish Gupta, Sayedali Mirjalili, and Yuantong Gu.
\newblock Comprehensive systematic review of information fusion methods in smart cities and urban environments.
\newblock {\em Information Fusion}, page 102317, 2024.

\bibitem{razi2023deep}
Abolfazl Razi, Xiwen Chen, Huayu Li, Hao Wang, Brendan Russo, Yan Chen, and Hongbin Yu.
\newblock Deep learning serves traffic safety analysis: A forward-looking review.
\newblock {\em IET Intelligent Transport Systems}, 17(1):22--71, 2023.

\bibitem{bastola2024driving}
Ashish Bastola, Julian Brinkley, Hao Wang, and Abolfazl Razi.
\newblock Driving towards inclusion: Revisiting in-vehicle interaction in autonomous vehicles.
\newblock {\em arXiv preprint arXiv:2401.14571}, 2024.

\bibitem{wang2022fast}
Hao Wang, Xiwen Chen, Abolfazl Razi, and Rahul Amin.
\newblock Fast key points detection and matching for tree-structured images.
\newblock In {\em 2022 International Conference on Computational Science and Computational Intelligence (CSCI)}, pages 1381--1387. IEEE, 2022.

\bibitem{rahmani2023graph}
Saeed Rahmani, Asiye Baghbani, Nizar Bouguila, and Zachary Patterson.
\newblock Graph neural networks for intelligent transportation systems: A survey.
\newblock {\em IEEE Transactions on Intelligent Transportation Systems}, 2023.

\bibitem{chen2022network}
Xiwen Chen, Hao Wang, Abolfazl Razi, Brendan Russo, Jason Pacheco, John Roberts, Jeffrey Wishart, Larry Head, and Alonso~Granados Baca.
\newblock Network-level safety metrics for overall traffic safety assessment: A case study.
\newblock {\em IEEE Access}, 11:17755--17778, 2022.

\bibitem{huang2023tap}
Baixiang Huang, Bryan Hooi, and Kai Shu.
\newblock Tap: A comprehensive data repository for traffic accident prediction in road networks.
\newblock In {\em Proceedings of the 31st ACM International Conference on Advances in Geographic Information Systems}, pages 1--4, 2023.

\bibitem{kipf2016gcnconv}
Thomas~N Kipf and Max Welling.
\newblock Semi-supervised classification with graph convolutional networks.
\newblock {\em arXiv preprint arXiv:1609.02907}, 2016.

\bibitem{defferrard2016chebnet}
Micha{\"e}l Defferrard, Xavier Bresson, and Pierre Vandergheynst.
\newblock Convolutional neural networks on graphs with fast localized spectral filtering.
\newblock {\em Advances in neural information processing systems}, 29:3844--3852, 2016.

\bibitem{hamilton2017inductive}
William~L Hamilton, Rex Ying, and Jure Leskovec.
\newblock Inductive representation learning on large graphs.
\newblock In {\em Proceedings of the 31st International Conference on Neural Information Processing Systems}, pages 1025--1035, 2017.

\bibitem{shi2021graphtransformer}
Yunsheng Shi, Zhengjie Huang, Shikun Feng, Hui Zhong, Wenjin Wang, and Yu~Sun.
\newblock Masked label prediction: Unified message passing model for semi-supervised classification, 2021.

\bibitem{zhang2023fptn}
Junhao Zhang, Juncheng Jin, Junjie Tang, and Zehui Qu.
\newblock Fptn: Fast pure transformer network for traffic flow forecasting.
\newblock In {\em International Conference on Artificial Neural Networks}, pages 382--393. Springer, 2023.

\bibitem{zhao2023dynamic}
Yusheng Zhao, Xiao Luo, Wei Ju, Chong Chen, Xian-Sheng Hua, and Ming Zhang.
\newblock Dynamic hypergraph structure learning for traffic flow forecasting.
\newblock In {\em 2023 IEEE 39th International Conference on Data Engineering (ICDE)}, pages 2303--2316. IEEE, 2023.

\bibitem{lan2022dstagnn}
Shiyong Lan, Yitong Ma, Weikang Huang, Wenwu Wang, Hongyu Yang, and Pyang Li.
\newblock Dstagnn: Dynamic spatial-temporal aware graph neural network for traffic flow forecasting.
\newblock In {\em International conference on machine learning}, pages 11906--11917. PMLR, 2022.

\bibitem{chen2022bidirectional}
Changlu Chen, Yanbin Liu, Ling Chen, and Chengqi Zhang.
\newblock Bidirectional spatial-temporal adaptive transformer for urban traffic flow forecasting.
\newblock {\em IEEE Transactions on Neural Networks and Learning Systems}, 34(10):6913--6925, 2022.

\bibitem{10580972}
Xiwen Chen, Huayu Li, Rahul Amin, and Abolfazl Razi.
\newblock Learning on bandwidth constrained multi-source data with mimo-inspired dpp map inference.
\newblock {\em IEEE Transactions on Machine Learning in Communications and Networking}, 2:1341--1356, 2024.

\bibitem{10621553}
Ashish Bastola, Hao Wang, Xiwen Chen, and Abolfazl Razi.
\newblock { FedMIL: Federated-Multiple Instance Learning for Video Analysis with Optimized DPP Scheduling }.
\newblock In {\em 2024 20th DCOSS-IoT}, pages 109--116.

\bibitem{watts1998collective}
Duncan~J Watts and Steven~H Strogatz.
\newblock Collective dynamics of ‘small-world’networks.
\newblock {\em nature}, 393(6684):440--442, 1998.

\bibitem{boccaletti2014structure}
Stefano Boccaletti, Ginestra Bianconi, Regino Criado, Charo~I Del~Genio, Jes{\'u}s G{\'o}mez-Gardenes, Miguel Romance, Irene Sendina-Nadal, Zhen Wang, and Massimiliano Zanin.
\newblock The structure and dynamics of multilayer networks.
\newblock {\em Physics reports}, 544(1):1--122, 2014.

\bibitem{barabasi1999emergence}
Albert-L{\'a}szl{\'o} Barab{\'a}si and R{\'e}ka Albert.
\newblock Emergence of scaling in random networks.
\newblock {\em science}, 286(5439):509--512, 1999.

\bibitem{ding2019application}
Rui Ding, Norsidah Ujang, Hussain~Bin Hamid, Mohd~Shahrudin Abd~Manan, Rong Li, Safwan Subhi~Mousa Albadareen, Ashkan Nochian, and Jianjun Wu.
\newblock Application of complex networks theory in urban traffic network researches.
\newblock {\em Networks and Spatial Economics}, 19(4):1281--1317, 2019.

\bibitem{bruna2013spectral}
Joan Bruna, Wojciech Zaremba, Arthur Szlam, and Yann LeCun.
\newblock Spectral networks and locally connected networks on graphs.
\newblock {\em arXiv preprint arXiv:1312.6203}, 2013.

\bibitem{kipf2016gcn}
Thomas~N Kipf and Max Welling.
\newblock Semi-supervised classification with graph convolutional networks.
\newblock {\em arXiv preprint arXiv:1609.02907}, 2016.

\bibitem{bianchi2021armanet}
Filippo~Maria Bianchi, Daniele Grattarola, Lorenzo Livi, and Cesare Alippi.
\newblock Graph neural networks with convolutional arma filters.
\newblock {\em IEEE Transactions on Pattern Analysis and Machine Intelligence}, page 1–1, 2021.

\bibitem{du2018tagcn}
Jian Du, Shanghang Zhang, Guanhang Wu, Jose M.~F. Moura, and Soummya Kar.
\newblock Topology adaptive graph convolutional networks, 2018.

\bibitem{velivckovic2017gat}
Petar Veli{\v{c}}kovi{\'c}, Guillem Cucurull, Arantxa Casanova, Adriana Romero, Pietro Lio, and Yoshua Bengio.
\newblock Graph attention networks.
\newblock {\em arXiv preprint arXiv:1710.10903}, 2017.

\bibitem{li2021spatial}
Mengzhang Li and Zhanxing Zhu.
\newblock Spatial-temporal fusion graph neural networks for traffic flow forecasting.
\newblock In {\em Proceedings of the AAAI conference on artificial intelligence}, volume~35, pages 4189--4196, 2021.

\bibitem{vaswani2017attention}
A~Vaswani.
\newblock Attention is all you need.
\newblock {\em Advances in Neural Information Processing Systems}, 2017.

\bibitem{dwivedi2023benchmarking}
Vijay~Prakash Dwivedi, Chaitanya~K Joshi, Anh~Tuan Luu, Thomas Laurent, Yoshua Bengio, and Xavier Bresson.
\newblock Benchmarking graph neural networks.
\newblock {\em Journal of Machine Learning Research}, 24(43):1--48, 2023.

\bibitem{zhou2023Spatiotemporal}
Zhengyang Zhou, Yang Wang, Xike Xie, Lianliang Chen, and Chaochao Zhu.
\newblock Foresee urban sparse traffic accidents: A spatiotemporal multi-granularity perspective.
\newblock {\em IEEE Transactions on Knowledge and Data Engineering}, 34(8):3786--3799, 2022.

\bibitem{prince2023understanding}
Simon~J.D. Prince.
\newblock {\em Understanding Deep Learning}.
\newblock The MIT Press, 2023.

\bibitem{xu2018gnn}
Keyulu Xu, Weihua Hu, Jure Leskovec, and Stefanie Jegelka.
\newblock How powerful are graph neural networks?
\newblock {\em arXiv preprint arXiv:1810.00826}, 2018.

\bibitem{gilmer2017mpnn}
Justin Gilmer, Samuel~S Schoenholz, Patrick~F Riley, Oriol Vinyals, and George~E Dahl.
\newblock Neural message passing for quantum chemistry.
\newblock In {\em International conference on machine learning}, pages 1263--1272. PMLR, 2017.

\bibitem{xie2018cgc}
Tian Xie and Jeffrey~C. Grossman.
\newblock Crystal graph convolutional neural networks for an accurate and interpretable prediction of material properties.
\newblock {\em Phys. Rev. Lett.}, 120:145301, Apr 2018.

\bibitem{li2020gen}
Guohao Li, Chenxin Xiong, Ali Thabet, and Bernard Ghanem.
\newblock Deepergcn: All you need to train deeper gcns, 2020.

\end{thebibliography}
